\documentclass{article}
\usepackage{spconf,amsmath,epsfig}
\usepackage{xcolor}
\usepackage{url}
\usepackage{booktabs}

\let\OLDthebibliography\thebibliography
\renewcommand\thebibliography[1]{
  \OLDthebibliography{#1}
  \setlength{\parskip}{0pt}
  \setlength{\itemsep}{0pt plus 0.3ex}
}

\begin{document}\sloppy

\def\x{{\mathbf x}}
\def\L{{\cal L}}

\title{Deep Tiered Image Segmentation for \\ Detecting Internal Ice Layers in Radar Imagery}
%

\name{Yuchen Wang$^1$ \; Mingze Xu$^1$ \; John D. Paden$^2$ \; Lora S. Koenig$^3$ \; Geoffrey C. Fox$^1$ \; David J. Crandall$^1$}
\address{$^1$Indiana University \;\; $^2$University of Kansas \;\; $^3$University of Colorado}

\newcommand{\xhdr}[1]{\vspace{5pt} \noindent {\textbf{#1}}}
\newcommand{\etal}[1]{\textit{et al}.}
\newcommand{\ie}[1]{\textit{i.e}.}
\newcommand{\quotes}[1]{``#1''}
\newcommand{\mingze}[1]{{\textcolor{brown}{Mingze: #1}}}
\newcommand{\note}[1]{{\textcolor{red}{ #1 }}}

\maketitle

\begin{abstract}
Understanding the structure of Earth's polar ice sheets is important
for modeling how global warming will impact polar ice and, in turn,
the Earth's climate.  Ground-penetrating radar is able to collect
observations of the internal structure of snow and ice, but the
process of manually labeling these observations is slow and laborious.
Recent work has developed automatic techniques for finding the
boundaries between the ice and the bedrock, but finding internal
layers -- the subtle boundaries that indicate where one year's ice
accumulation ended and the next began -- is much more challenging
because the number of layers varies and the boundaries often merge and
split.  In this paper, we propose a novel deep neural network for
solving a general class of tiered segmentation problems.  We then
apply it to detecting internal layers in polar ice, evaluating on a
large-scale dataset of polar ice radar data with human-labeled
annotations as ground truth.
\end{abstract}
%
\section{Introduction}

Understanding the impacts of global climate change begins at the north
and south poles: as the earth warms and the vast polar ice breaks
apart and melts, sea levels will rise and absorb more solar energy,
which in turn will cause the Earth to warm even faster~\cite{climate}.
%
%
To predict and potentially mitigate these changes,
glaciologists have developed models of how polar ice and snow will
react to changing climates.  But these models require detailed
information about the current state of the ice. While we may think of
polar ice sheets as simply vast quantities of frozen water, in reality
they have important structure that influences how they will react to
rising temperatures. For example, deep beneath the ice is 
bedrock, which has all the same diverse features as the rest of the
Earth's surface -- mountains, valleys, ridges, etc.~-- that affect how melting ice will behave. The ice
sheets themselves also have structure: snow and ice accumulate in
annual layers year after year, and these layers record important
information about past climatological events that can help predict the
future.

\begin{figure}
    \begin{center}
        \includegraphics[width=0.48\textwidth]{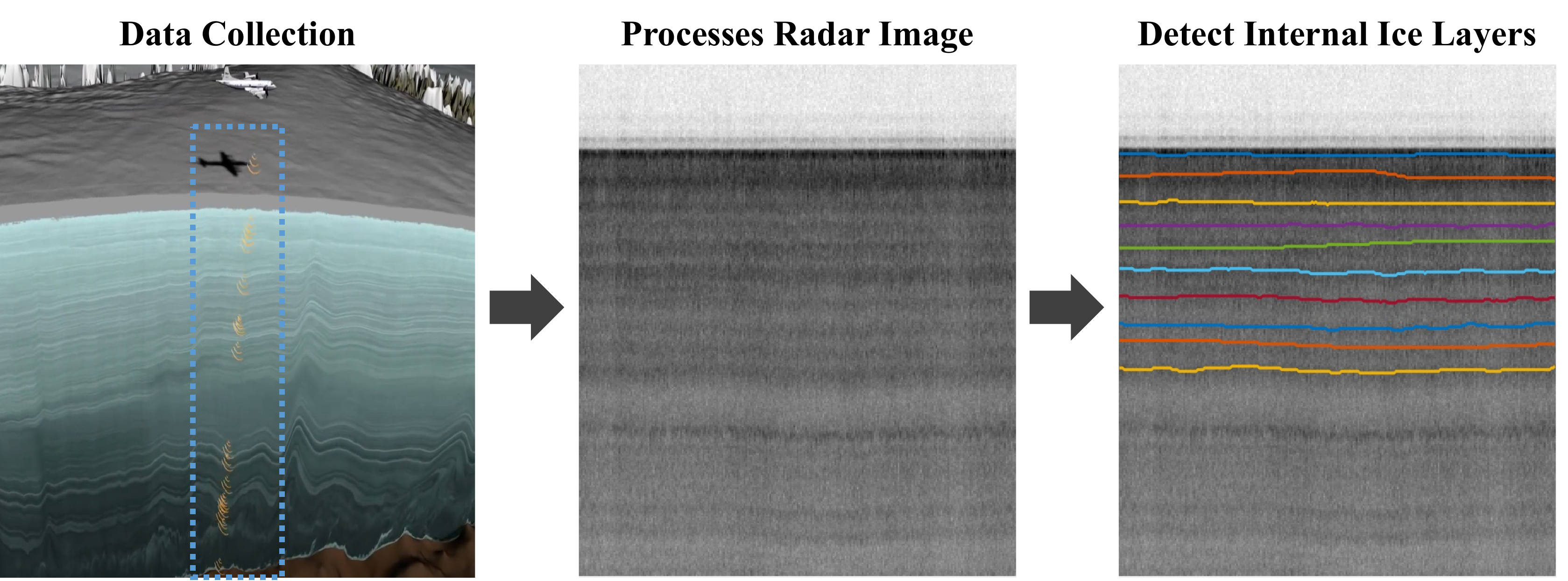}
    \end{center}
    \vspace{-15pt}
    \caption{Given an echogram from ground-penetrating radar over polar ice, 
    we automatically estimate the number and position of internal ice layers. Left figure is adapted from \cite{nasayoutube}.}
    \label{fig:teaser}
    \vspace{-5pt}
\end{figure}

To directly collect data about the structure of ice requires drilling
ice cores, which is a slow, expensive, and extremely laborious
process. Fortunately, ground-penetrating radar systems have been developed that can
fly above the ice  and collect data about the material
boundaries deep under the surface. This
process generates radar echograms
(Fig.~\ref{fig:teaser}), where the vertical axis represents the depth
of the return, the horizontal axis corresponds to distance along the
flight path, and the pixel brightness indicates the amount of energy
scattered from the subsurface structure. However, the echograms are
very noisy and typically require laborious  manual
annotation~\cite{macgregor}.

Most automatic techniques for finding layers in these images
only consider the ice-air and
ice-bedrock boundaries, which are the most prominent~\cite{crandall2012layer,lee2014estimating,xu2017automatic,xu2018multi,kamangir2018deep}.
 A much more challenging problem is to identify the
``internal'' layers of the ice and snow caused by annual accumulation.
At first glance, solving this problem may seem like a straightforward application of
traditional computer vision techniques.
However, approaches based on edge detection do not work well
because 
the layer boundaries are subtle, the noise characteristics vary dramatically, and the 
number of visible layers changes across
different regions of ice.
Unlike most segmentation problems in computer vision,
the layers
here do not correspond to
``objects'' with distinctive colors or textures.




Nevertheless, our problem
can be viewed as a generalization of the tiered scene segmentation
problem~\cite{tiered}. Tiered segmentation partitions an image
into a set of regions $\{r_1, r_2, ..., r_n\}$ such that in each image
column, all pixels belonging to $r_i$ are above (have lower row index
than) all pixels corresponding to $r_j$ for $i<j$. Felzenswalb and
Veksler~\cite{tiered} solved this problem using energy minimization
with dynamic programming, but they assumed no more than three distinct
labels per column because their inference time was exponential in the
number of labels.

In this paper, we revisit tiered labeling using deep learning,
and we consider a more challenging problem in which the number of labels is large and unknown ahead of time.
We propose a novel deep neural network
which performs the tiered segmentation in two stages. We first use
a 2D convolutional network (CNN) to simultaneously solve three problems:
detecting the position of the top layer, roughly estimating the
average layer thickness, and estimating 
the number of visible layers.
Propagating the estimated first layer downward using the estimated thickness gives a rough approximation
of the tiered segmentation.
Then we refine the pixel-level boundary positions 
using a recurrent neural network (RNN) to account for differences across
different layers.
We evaluate our method on
internal ice layer segmentation on a large-scale, publicly-available
polar echogram dataset. 
Experimental results show that our approach significantly outperforms
baseline methods, and is especially efficient on multi-layer
detection.
Beyond polar ice, our technique is
general and can be applied to tiered segmentation 
problems in other domains.

\section{Related Work}

Crandall~\etal \quad ~\cite{crandall2012layer} detected two specific
types of layer boundaries (ice-air and ice-bed) in echograms using
discrete energy minimization with a pretrained template model and a
smoothness prior.  Lee~\etal \quad ~\cite{lee2014estimating} proposed
a more accurate method using Gibbs sampling from a
joint distribution over all candidate layers, while Carrer and
Bruzzone~\cite{carrer2017automatic} further reduced the computational
cost with a divide-and-conquer strategy.  Xu~\etal \quad
~\cite{xu2017automatic} extended the work to estimate 3D ice surfaces
using a Markov Random Field (MRF), and Berger~\etal \quad
~\cite{berger2018automated} followed up with better cost functions
that incorporate domain-specific priors.
More recent work has applied deep learning. Kamangir~\etal \quad ~\cite{kamangir2018deep}
detected ice boundaries using convolutional
neural networks applied to wavelet features. Xu~\etal \quad ~\cite{xu2018multi} proposed a
multi-task spatiotemporal neural network to reconstruct 3D ice
surfaces from sequences of tomographic images.  However, all of this
work focuses on detecting a small, known number of layer boundaries
(typically two) and thus is not appropriate for internal layers, because
the number of visible internal layers varies and may be quite large.

Very recent work, contemporaneous to ours, has considered the internal layer
detection problem.
Varshney~\etal \quad~\cite{varshney2020deep}
treat the problem as semantic segmentation, while
Yari~\etal \quad~\cite{yari2020multi} classify pixels into layer boundaries or not,
which is a binary classification problem. Those papers require
post-processing steps either to smooth the inconsistent labels between
layers or to specify the layer indices.

Our problem can be thought of as a more general version of 
the tiered segmentation problem~\cite{tiered} proposed by Felzenswalb
and Veksler, who presented an algorithm based on dynamic programming.
However, their solution required the number of
tiers (labels) to be fixed ahead of time to a small number (3) because
their inference was exact and {exponential in the number of labels},
and used hand-crafted features.
In this paper, we propose a new approach to a more general version of the
tiered segmentation problem, in which the number of labels can be large
and unknown. Our technique combines convolutional
and recurrent neural networks for counting and detecting
an arbitrary number of layer boundaries.

\section{Methodology}

\begin{figure*}
    \begin{center}
        \includegraphics[width=0.90\textwidth]{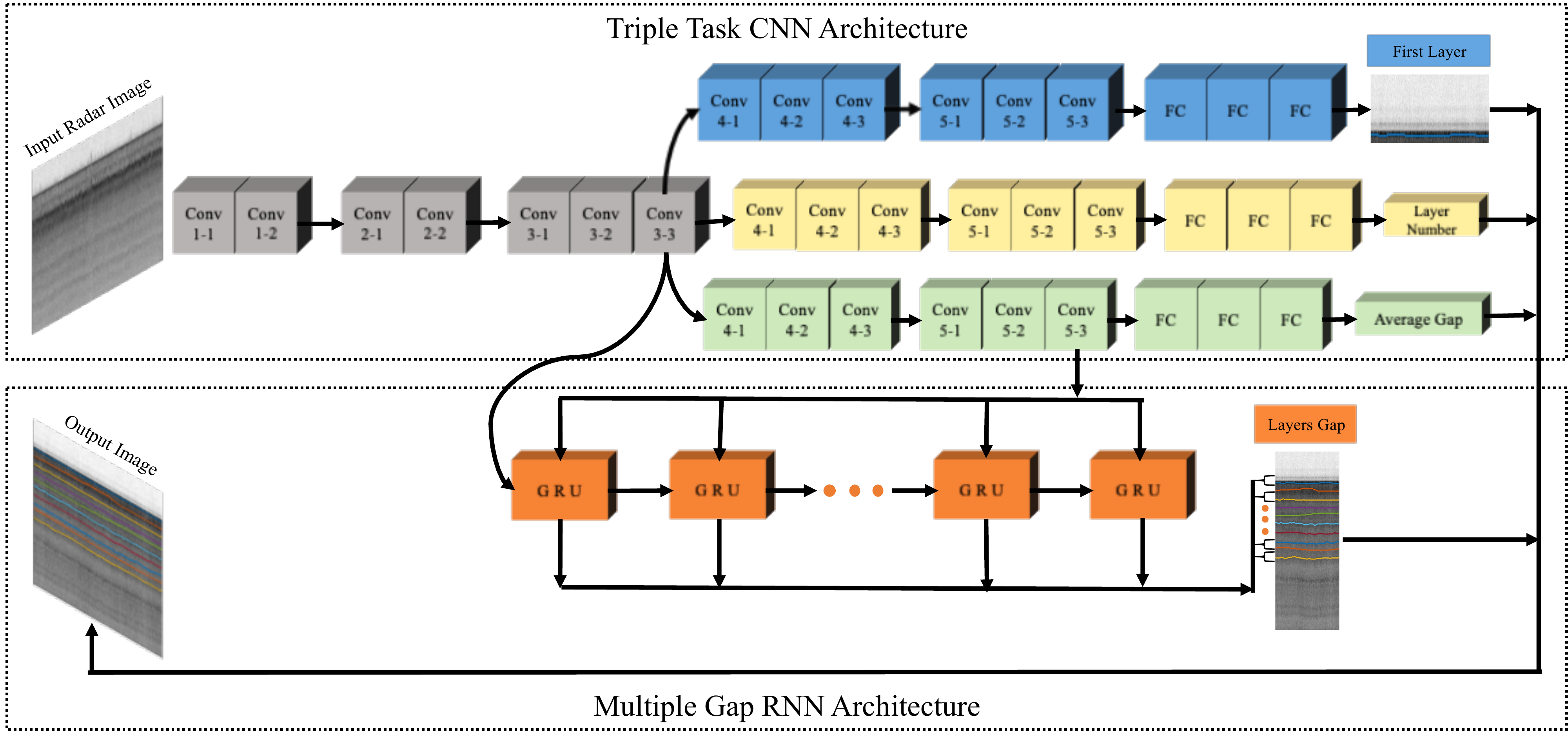}
    \end{center}
    \vspace{-15pt}
    \caption{Architecture of our model for detecting internal ice
      layers in radar echogram images. Through a combination of CNN and RNN
      networks, we estimate both the number of layers and the boundaries of each layer.}
    \label{fig:model}
    \vspace{-5pt}
\end{figure*}

Given a noisy radar echogram $I$, which is a 2D image of size $1\times H \times W$ pixels,
our goal is to localize $N$ internal ice layer boundaries
and exactly one surface boundary between the ice and air. The
output thus should be $N + 1$ boundaries. We need to estimate both the
number of boundaries $N$ (which varies from image to image, although our
implementation assumes $N<30$) and all the boundary locations
based on noisy and ambiguous data.

Our technique encodes the physical constraints of this tiered
segmentation problem. First, since the labeled regions correspond to
physical layers, layer boundaries cannot cross; more precisely, we
partition the image into regions $\{r_1, r_2, ..., r_n\}$ such that in
each image column, all pixels belonging to $r_i$ are above all pixels
corresponding to $r_j$ for $i<j$. Second, we assume that adjacent
boundaries are roughly parallel, which is reasonable since the amount of
snow or ice that falls in any given year is roughly consistent across
local spatial locations. Finally, we assume that the thickness of
different layers is roughly the same at any given spatial location,
which is reasonable since the amount of snow or ice is similar across
different years. These are all rough, weak assumptions, and our model
is able to handle the significant deviations from them that occur in
real radar data.

We address this problem using two main steps, following the intuition
that a human annotator might use:  first do a rough
top-down segmentation of the image to incorporate global constraints
on the layer structure, and then use that rough segmention to do a
bottom-up refinement of the layer boundaries.  More specifically, we
first design a triple-task Convolutional Neural Network (CNN) model to
estimate the number of ice layers $\hat{N}$, the location of the top
layer $\hat{F}$ (encoded as a $W$-d vector indicating the row index
for each column of the image), and the average thickness of all layers 
(the average vertical gap between boundaries) in the echogram.
The top boundary is
typically quite prominent since it is between air and ice, and
provides a strong prior on the shape of the much weaker boundaries
below.  Second, we design a Recurrent Neural Network (RNN) to estimate
$\hat{GapM}$, an $N \times W$ matrix encoding the thickness (gap
between adjacent boundaries) for each layer at each column, based on
the estimates in the first step.

To generate the final segmentation, we combine $\hat{F}$ and $\hat{GapM}$ according to $\hat{N}$ to generate
output $\hat{M}$ which is a $(N+1) \times W$ matrix. Each element in
$\hat{M}$ indicates the row index for a given boundary at a given column in the
input $I$.

\subsection{Triple Task CNN (CNN3B)}

Our first step applies a three-branch Convolutional Neural Network
(CNN) to roughly estimate the surface boundary location, the number of
layer boundaries, and the average thickness of each internal layer across the
echogram.
 Fig.~\ref{fig:model} shows our CNN architecture, which was
inspired by Xu~\etal \quad ~\cite{xu2018multi} but with significant
modifications. Our model takes a 2D image $I$ as input. Then we use
three shared convolutional blocks, each of which is followed by max
pooling operations. The shared convolutional blocks are used to
extract low-level features for the next three branches, because
similar evidence is useful for estimating the first layer, the number
of layers, and the average thickness.

The model then divides into three branches.
The first branch estimates the position of the surface layer, and
uses six convolutional layers for
modeling features specific to the first layer and one fully connected
layer to generate outputs
$\hat{F}=\{ \hat{f_1},\hat{f_2},\cdots,\hat{f_W} \}$. Each element
represents the row coordinate of the first layer within that
column. The ground truth vector $F = \{ f_1,f_2, \cdots,f_W \}$ is generated
from the top boundary of the human-labeled ground truth $M_N =
\{ m_{N,1},m_{N,2}, \cdots,m_{N,W} \}$. The loss function for estimating the first layer uses an L1 Manhattan distance to
encourage the model's output to agree with
the human-labeled ground truth,
\begin{equation}
L_{fl}=\frac{1}{W} \sum_{w=1}^{W}\left|\hat{f}_{w}-f_{w}\right|.
\end{equation}
The second branch predicts the number of ice layer boundarise, and
includes six convolutional layers and three fully connected layers.
We view this as a classification problem, so this 
branch produces a vector $v$ which is a probability distribution over a
discrete set of possible numbers of boundaries. In our experiments, we assume $N < 30$, so $v$ is 31-dimensional.
The ground truth
is the number of labeled boundaries $N$ in the human-annotated ground truth of the image. Cross-entropy loss
is used during training,
\begin{equation}
L_{number}=-\log \left(\frac{\exp (v[\operatorname{N}])}{\sum_{j} \exp (v[j])}\right).
\end{equation}
The third branch roughly estimates the average thickness (gap) of all the layers in the echogram,
and follows the same general design as the first branch but with a
single scalar output from the final fully connected layer. The loss
calculates the absolute value between the output $\hat{\Delta}$
and the ground truth $\Delta$,
\begin{equation}
L_{\Delta}=\left|\hat{\Delta}-\Delta\right|.
\end{equation}
Finally, our CNN loss function combines the three branches,
\begin{equation}
L= L_{fl} + L_{number} + L_{\Delta}.
\end{equation}
We use VGG16~\cite{simonyan2014very}
as the network backbone.

\subsection{Multiple Gap RNN}

Having roughly estimated the global structure of the echogram in the last section,
we next use
an RNN 
to generate a more accurate gap (thickness) value for each layer  in the echogram.
We use Gated Recurrent
Units (GRUs)~\cite{chung2015gated}, which require less computational cost and
are easier to train than LSTMs~\cite{hochreiter1997long}.

As shown in Fig.~\ref{fig:model}, our model has one hidden layer, wherein
each GRU cell takes feature map $Avg_F$
generated before the fully connected layer of our Triple Task CNN
model's third branch and the output of the previous GRU cell as
inputs, and produces $W$ real-valued numbers indicating the predicted
gap between layer boundaries within each column of the data. $Avg_F$ is
projected to the size of the GRU hidden state with a fully connected
layer before GRU takes it as input. During training, the GRU cell
is operated for $N$ iterations, where each iteration $n$ estimates the
gap between layer $n+1$ and layer $n$. In a given iteration $n$,
the GRU cell takes the projected $Avg_F$ as input. The GRU cell
outputs a sequence of hidden states $\{ h_1,h_1, \cdots,h_n\}$ with
iteration $n \in [1,N]$, and each hidden state $h_n$ is followed by a
fully-connected layer to predict gap value $\hat{GapM_n}$. We use L1
Manhattan distance to supervise the model to predict $\hat{GapM}$
according to the human-labeled ground truth $GapM$,
%
\begin{equation}
  L_{GapM}=\frac{1}{N}\frac{1}{W}\sum_{n=1}^{N}\sum_{w=1}^{W}
  \left|\hat{GapM_{n,w}}-GapM_{n,w}\right|.
\end{equation}
\vspace{-15pt}
\subsection{Combination}
We combine our Triple Task CNN and Multiple Gap RNN to predict the
number of internal ice layer boundaries and their positions in the input image
$I$. The RNN uses general features as shown in Fig.~\ref{fig:model} to
initialize the GRU's hidden state and takes an average feature map
$Avg_F$ as input.  Based on the first layer output and the number of
layers from the Triple Task CNN, our model generates the first boundary
$M_0$ ($W$-d vector) in our result $\hat{M}$. We then apply the layer gap output
$GapM$ predicted by our multiple Gap RNN according to the first layer result,
\begin{equation}
\begin{array}{c}
M_{i}=M_{i-1}+GapM_{i}, i \in(1, N),
\end{array}
\end{equation}
where N is number of layer boundaries.
In addition, $M_{i}$ and $GapM_{i}$ are both
$W$-d vectors. For each image, we compute all $M_i$'s to create
$\hat{M}$ which is a $(N+1,W)$ matrix, and compare it with ground
truth $M$ to evaluate our model.

\section{Experiments}

\vspace{-5pt}
\subsection{Dataset}
We use the annual ice layer dataset collected by the
Center for Remote Sensing of Ice Sheets (CReSIS) at the University of Kansas
and the National Snow and Ice Data
Center at the University of Colorado~\cite{koenig2016annual}.
The data is collected by ultra-wideband snow radar 
operated over a frequency range from 2.0 to 6.5 GHZ, and
consists of 17,529 radar images
with human-labeled annotations that identify the positions of 
internal ice layers. Formally, our task is to detect all internal ice
layers $\hat{M}$ in a given single-channel image $I$. Each 
element in $\hat{M}$ indicates the 
row coordinate (in the range $[1, H]$, where $H$ is image height) of an ice layer
for a given column.

\xhdr{Preprocessing.}
We resize all input images to
$300 \times 256$ by using bi-cubic interpolation. We normalize 
the grayscale pixel values by
subtracting the mean and dividing by the standard deviation (both 
of which are calculated from the training
data). Following~\cite{xu2018multi}, we also normalize the ground truth
row labels to a coordinate system spanning $[-1, 1]$ in each image.
We also remove input images that have missing data.

\vspace{-5pt}
\subsection{Implementation Details}
We use PyTorch to implement our model and do the
training and all experiments on a system with Pascal Nvidia
Titan X graphics card.
We randomly choose 80\% of images for training and 20\% for testing.
The Adam optimizer with default parameters is used to learn the CNN
parameters with batch size of 16. The training process is stopped
after 30 epochs, starting with a learning rate of $10^{-4}$ and
reducing in half every 10 epochs. The RNN training uses the same
optimizer, update rule, and batch size as the CNN's,
but initial learning rate is set to $10^{-3}$.

\subsection{Evaluation Metrics}

Prior work has used mean absolute error in pixels between predicted and
ground truth layers~\cite{crandall2012layer,lee2014estimating}, a familiar
evaluation metric in signal processing applications. However, in our
problem of internal ice detection, the number of layers is unknown,
which means the evaluation metric must capture both the accuracy of
estimated layer count and the localization accuracy of the layers.
Prior work on internal layer detection
has typically been evaluated qualitatively~\cite{mitchell2013semi}.

We thus introduce two quantitative, objective evaluation approaches for the
tiered segmentation problem.
Our first evaluation protocol assumes that the correct number of layers is known
via an oracle, and then measures mean absolute error in pixels.
Assuming that the correct number of layers is given  is useful both for
isolating the accuracy of layer localization, and for allowing comparison with models that are not able
to estimate layer counts.
To evaluate the accuracy of both the layer count and layer boundaries,
we propose \textit{layer-AP} based on average precision. For each estimated layer, we search through
the ground truth layers to find the closest match according to mean absolute error.
Each ground truth layer is only allowed to match one estimated layer. 
Then we define a set of threshold values $t_l$. For each threshold, we count the number of estimated layers which have a 
mean absolute error in pixels under the threshold, and call this the number of matches $m_i$ for that threshold.
In particular, the layer average precision is computed as,
\begin{equation}
\textrm{layer-AP} = \frac{1}{l}\sum_{i=1}^{l} \frac{m_{i}}{N+1},
\end{equation}
where $N+1$ is the number of layer boundaries ($N$ is the number of gaps between boundaries), and $l$ indicates the number of
thresholds. In this work, we use 10 different thresholds and set $t_l = [1,4,7,10,13,16,19,22,25,27]$,
assuming input images are in size of $300 \times 256$.

\subsection{Baselines}
We are not aware of any existing work that solves
the problem we consider here:
existing fully-automatic approaches
to the tiered segmentation problem
assume the number of layers is no greater than two and is known ahead of time.
We thus develop some baseline models to compare our results
against.

Crandall~\etal \quad ~\cite{crandall2012layer} proposed a technique based on graphical models to find layer boundaries, which we call \textbf{Sequential}. However, they assume
exactly two layer boundaries because the running time is exponential in the number of layers. Here, we adapted it to our problem by
using an oracle to determine the number of layers (by looking at ground truth), and
then running sequentially to find each layer one-by-one.
\textbf{Naive CNN} uses the
VGG16~\cite{simonyan2014very} as backbone
which directly predicts a fixed number
of internal layers by producing a label matrix in one-shot.
\textbf{RNN30} models the dependencies in the vertical direction:
given the estimated boundary for a given layer and previous layers, it
predicts the boundary for the subsequent (next-deeper) annual layer.
\textbf{RNN256} is a baseline that uses a recurrent neural network (RNN) 
to model sequential dependencies across columns, assuming a fixed number
of layers.
\textbf{CNN2B} is a simpler version of our model that uses
only two branches, one to predict the top
layer (the air and ice boundary), and 
one to predict the average gap between layers.
\textbf{CNN3B} is a version of our model with all three branches, but without the RNN refinement. 
\textbf{CNN3B+RNN} is our full model described above.

\subsection{Evaluation Results}

\vspace{-5pt}
\xhdr{Quantitative results} are presented in Table~\ref{tab:mean_results} and~\ref{tab:lap_results}
in terms of mean absolute error and layer-AP, respectively.
In each table, we present two sets of results: one in which the number
of layers is known ahead of time by an oracle (\textit{i.e.}, by consulting
the ground truth), and one in which it is predicted automatically.
Note that only the techniques that use \textit{CNN3B} are able to estimate
the number of layers automatically, which is why the other results
are listed as missing in the table. For calculating mean absolute error
when models incorrectly estimate the number of layers, 
we pad either the ground truth or the output (whichever has fewer layers) with extra layers consisting of zero vectors to penalize
these incorrect estimations.

\begin{figure*}
    \begin{center}
        \includegraphics[width=17cm]{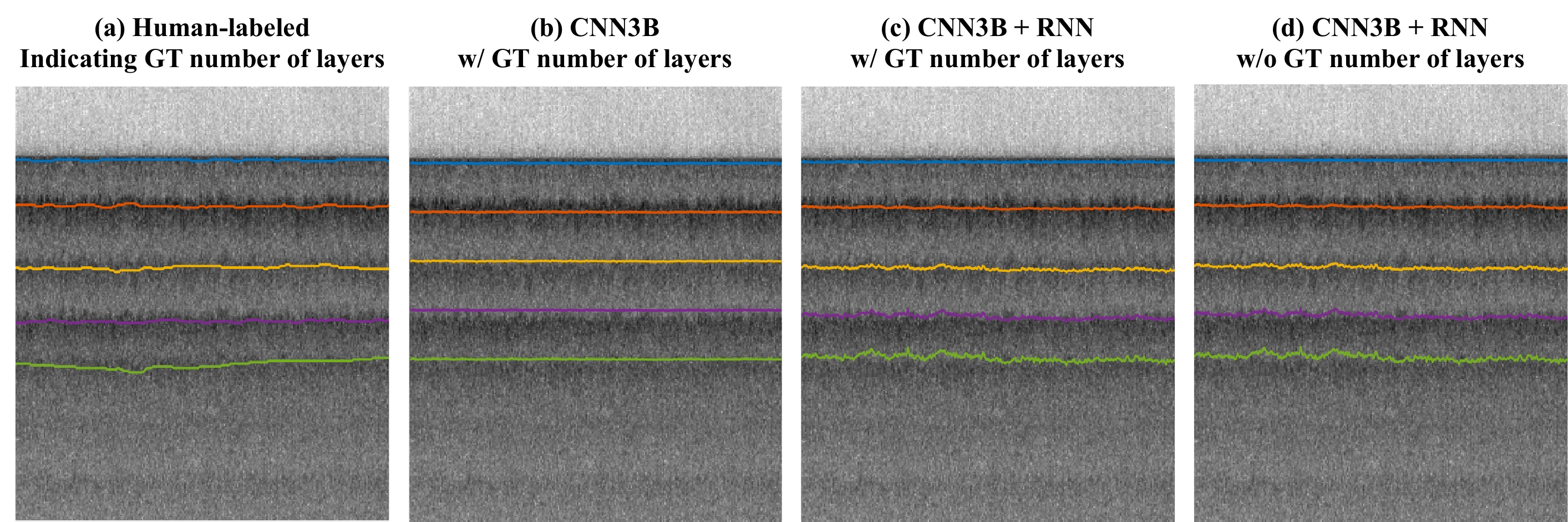}
    \end{center}
    \vspace{-20pt}
    \caption{Sample results, showing (a) ground truth, (b)
      \textit{CNN3B} output with ground truth number of layers, (c) \textit{CNN3B+RNN} with ground truth number of
      layers, and (d)   \textit{CNN3B+RNN} with estimated number of layers.}
    \label{fig:sample_results}
\end{figure*}


Comparing with other models in Table~\ref{tab:mean_results} and~\ref{tab:lap_results},
we observe that our combination of 
CNN3B and RNN models significantly outperforms all baselines
in terms of both mean average error and layer-AP. 
Our two models \textit{CNN3B} and \textit{CNN3B+RNN} have the ability
to estimate the number of internal ice layers, and reach $85.2\%$
accuracy on this layer counting task, which is why their accuracy
decreases only slightly when the number of layers is not provided by the oracle.
Our model \textit{CNN3B+RNN} shows the best results of all other
baselines, even when our model must estimate the number of layers and the 
the baselines know it from the oracle.

\begin{table}
  \begin{center}
\scalebox{0.88}{\begin{tabular}{lcc}
\toprule
            & \multicolumn{2}{c}{Mean Error (in pixels) $\downarrow$}                                       \\ \cmidrule{2-3} 
            & \multicolumn{1}{l}{\# layers from oracle} & \multicolumn{1}{l}{\# layers estimated} \\ \midrule
Sequential \cite{crandall2012layer}   &  88.98  & -  \\
Naive CNN   & 24.32  & -    \\
RNN30       & 21.79  & -    \\
RNN256      & 20.20  & -    \\
CNN2B       & 11.94  & -    \\
CNN3B       & 7.91   & 9.27 \\
CNN3B+RNN   & 6.96   & 8.73 \\ \bottomrule
\end{tabular}}
  \end{center}
  \vspace{-12pt}
\caption{Evaluation results by measuring the error in terms of the mean absolute column-wise difference compared to ground truth, in pixels.}
\label{tab:mean_results}
\end{table}

\begin{table}
  \begin{center}
\scalebox{0.88}{\begin{tabular}{lcc}
\toprule
            & \multicolumn{2}{c}{layer-AP  $\uparrow$}                                                     \\ \cmidrule{2-3} 
            & \multicolumn{1}{l}{\# layers from oracle} & \multicolumn{1}{l}{\# layers estimated} \\ \midrule
Sequential \cite{crandall2012layer}   & 0.059  & -  \\
Naive CNN   & 0.183  & -      \\
RNN30       & 0.218  & -      \\
RNN256      & 0.254  & -      \\
CNN2B       & 0.635  & -      \\
CNN3B       & 0.843  & 0.822  \\
CNN3B+RNN   & 0.882  & 0.853  \\ \bottomrule
\end{tabular}}
\end{center}
\vspace{-12pt}
\caption{Evaluation results by measuring the layer average precision with thresholds compared to ground truth.}
\vspace{-5pt}
\label{tab:lap_results}
\end{table}

\xhdr{Qualitative results} are shown in Fig.~\ref{fig:sample_results}.
The first column shows the human annotated layers, while the second
column shows the result generated by one of our baselines,
\textit{CNN3B}.  The results of this baseline roughly agree with the
ground truth, but all layers except the first
show different degrees of inaccurate localization compared with the human
annotations. The third and fourth columns show the
results with and without the layer number oracle. Since our
\textit{CNN3B+RNN} model is highly accurate at estimating the number of layers,
the output with and without the oracle are nearly the same.
We provide additional sample results in 
supplementary material.

As shown in the examples, our model only needs the input image
to generate results that are very close to human annotations in most cases.
The improvement between \textit{CNN3B} and \textit{CNN3B+RNN} indicates that the RNN
contributes to our final result even though the \textit{RNN30} and
\textit{RNN256} baselines fail to work well on their own. The results show that
both steps of our model are important to achieve high
performance. 

\vspace{-5pt}
\section{Conclusion}
\vspace{-2pt}

We have considered a generalization of the tiered segmentation problem
and apply it 
to a problem of great societal consequence: automatically
understanding the internal layer structure of the polar ice sheets
from ground-penetrating radar echograms.
We show that our approach can effectively estimate arbitrary numbers of snow or ice layers from noisy radar images.
Experimental results on a challenging, publicly-available dataset demonstrate
the significant improvements over existing methods. 

\vspace{-2pt}
\section{Acknowledgments}
\vspace{-2pt}
This work was supported in part by
the National Science Foundation (DIBBs 1443054, CAREER IIS-1253549),
and by the IU Office of the Vice Provost for Research,
the IU College of Arts and Sciences,
and the IU Luddy School of Informatics, Computing, and Engineering
through the Emerging Areas of
Research Project “Learning: Brains, Machines, and Children.”
We acknowledge the use of data from CReSIS with support from
the University of Kansas and Operation IceBridge (NNX16AH54G).

\vspace{-5pt}
\bibliographystyle{IEEEbib}
\bibliography{icme2021template}
\section{More Experiment Results}
\begin{figure*}
     \begin{center}
         \includegraphics[width=17cm]{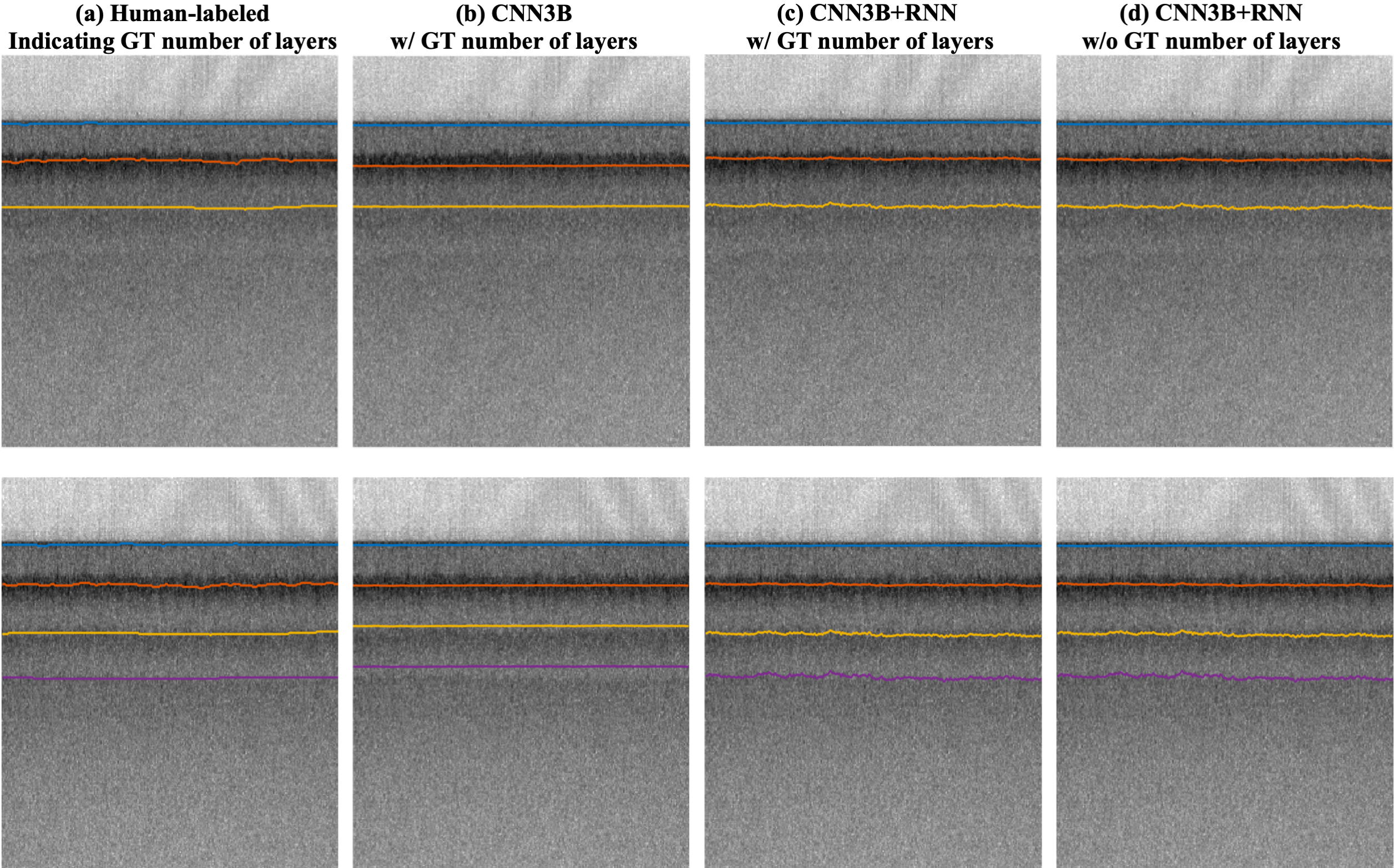}
     \end{center}
     \caption{Sample results. First column is input data with ground
       truth. Second column is one of our baselines, \textit{CNN3B},
       with ground truth number of layers. Third column is our
       best model, \textit{CNN3B+RNN}, with ground truth number of
       layers. Fourth column is our best model, \textit{CNN3B+RNN},
       with estimated number of layers.}
     \label{fig:sample_results_two}
 \end{figure*}

{Fig.~\ref{fig:sample_results_two}} shows two more examples with different numbers of
internal layers. In each example, the first column shows
the human annotated layers with input data as background.
The second column shows the estimated result generated by
one of our baselines, CNN3B. The results of this baseline
roughly agree with the ground truth, but have some clear
defects. For example, in the first row, the second orange
layer in the CNN3B result fails to match the ground truth
well. In the second row, the third yellow layer and fourth
purple layer show clear mismatches. 
The third and fourth columns show the prediction result
with and without the layer number oracle. Since our
CNN3B+RNN model successfully predicts the number of
layers in these three cases, the output with and without the
oracle are nearly the same. Both our CNN3B+RNN results
show improvements in both cases. In the first row, the
second orange layer matches the human annotation in the
first column better than CNN3B in the second column. In
the second row, the third yellow layer and fourth purple
layer are clearly closer to the ground truth than the result.

\begin{figure*}
     \begin{center}
         \includegraphics[width=17cm]{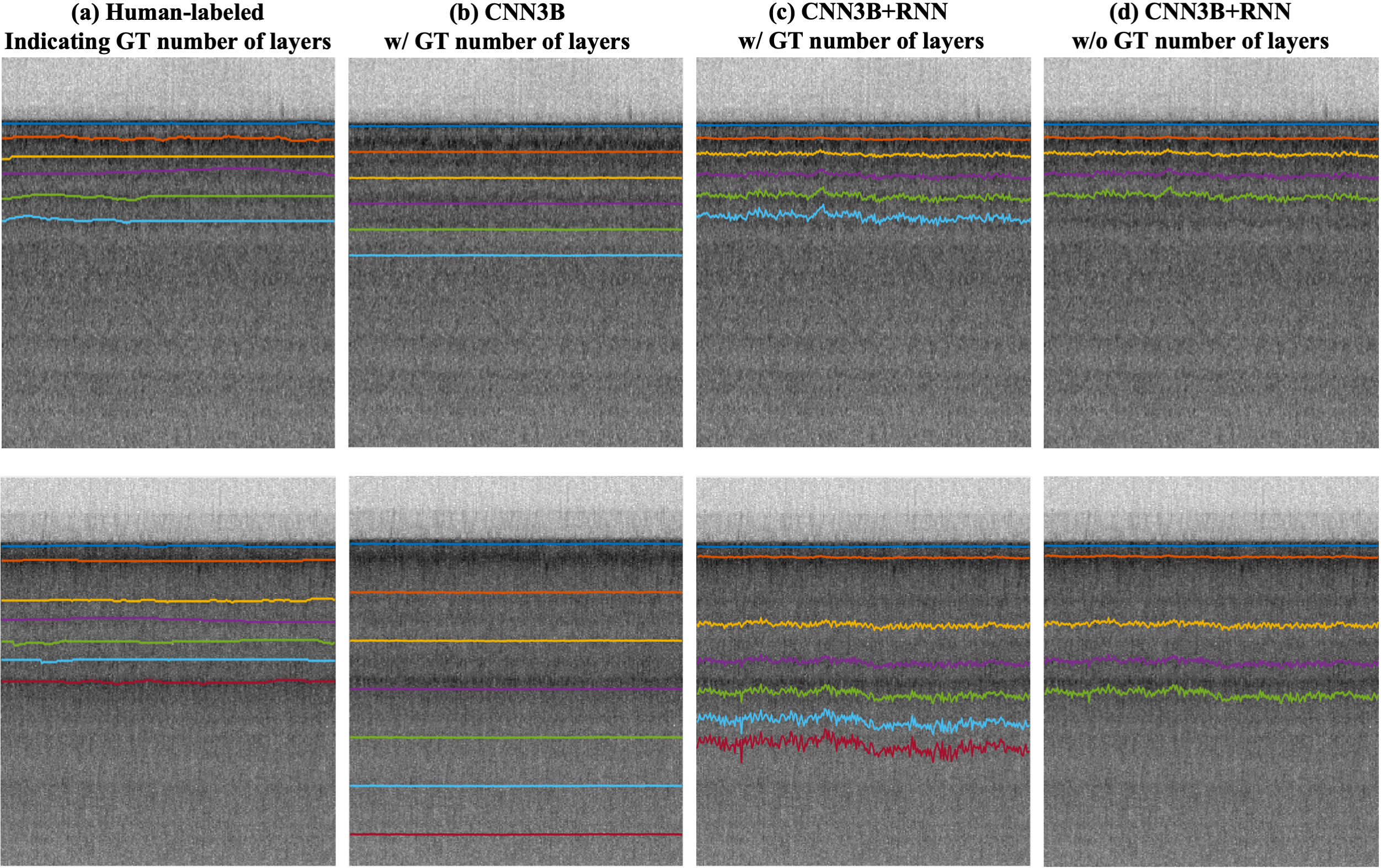}
     \end{center}
     \caption{Failure cases. First column is input data with ground truth. Second column is one of our baseline \textit{CNN3B} result with ground truth number of layers. Third column is our best model \textit{CNN3B+RNN} result with ground truth number of layers. Fourth column is our best model \textit{CNN3B+RNN} result with prediction number of layers. Two rows represent two different input data with annotations and prediction results.}
     \label{fig:failure_cases}
 \end{figure*}

There are two failure cases showed in Fig.~\ref{fig:failure_cases}.
 In the fist failure case, there are 6 internal layers in the ground truth, but \textit{CNN3B+RNN} without the oracle
 fails to predict the number of layers correctly. Our \textit{CNN3B} only predicts
 the first layer well and shows a clear mismatch in all the other
 layers, but our \textit{CNN3B+RNN} result in the fourth column predicts the 
 first 5 layers reasonably  well. Our \textit{CNN3B+RNN} with the oracle shows
 the best results for this case, but there are still many ripples in the result showing that our model failed to predict it perfectly.
 In the second failure case, our \textit{CNN3B} model fails to predict almost all the layers, while \textit{CNN3B+RNN} both fails to estimate the number of layers or match most of the internal layers. However,
 \textit{CNN3B+RNN} still works well for the first two layers.
 
 There are two causes for these failure cases. First, the input data is noisy and complex, and difficult to be annotated even by an experienced human annotator.
Not only does this means that our model must learn to process the complex and 
noisy input data, but it also means that the ``ground truth'' from human 
annotators also has much noise and inconsistency. This annotation noise
not only adds noise during training, but also means that we are measuring
error against ground truth that in itself has many errors.
Second, we face a significant unbalanced dataset issue: there only 1.35\% of the images have more than 5 internal ice layers, for example, 
which may affect our model's learning capability.

\end{document}